\documentclass[10pt,twocolumn,letterpaper]{article}

\usepackage{iccv}              %
\usepackage{xcolor,colortbl}

\definecolor{iccvblue}{rgb}{0.21,0.49,0.74}
\usepackage[pagebackref,breaklinks,colorlinks,allcolors=iccvblue]{hyperref}
\usepackage{subcaption}
\usepackage{multirow}
\usepackage{xcolor}
\definecolor{emerald}{HTML}{50C878}
\usepackage[accsupp]{axessibility}  %

\def\paperID{3350} %
\def\confName{ICCV}
\def\confYear{2025}

\title{ShortV: Efficient Multimodal Large Language Models by Freezing Visual Tokens in Ineffective Layers}

\author{
  \textbf{Qianhao Yuan}${}^{1,2}$, 
  \textbf{Qingyu Zhang}${}^{1,2}$, 
  \textbf{Yanjiang Liu}${}^{1,2}$, 
  \textbf{Jiawei Chen}${}^{1,2}$, \\
  \textbf{Yaojie Lu}${}^{1,}\thanks{Corresponding authors.}$, 
  \textbf{Hongyu Lin}${}^{1}$,
  \textbf{Jia Zheng}${}^{1,*}$,
  \textbf{Xianpei Han${}^{1}$},
  \textbf{Le Sun${}^{1}$}
  \\
  ${}^{1}$Institute of Software, Chinese Academy of Sciences\\
  ${}^{2}$University of Chinese Academy of Sciences \\
 \texttt{\{yuanqianhao2024,zhangqingyu2024,liuyanjiang2021,chenjiawei2024,} \\
 \texttt{luyaojie,hongyu,zhengjia,xianpei,sunle\}@iscas.ac.cn} \\
}

\begin{document}

\maketitle
\begin{abstract}

Multimodal Large Language Models (MLLMs) suffer from high computational costs due to their massive size and the large number of visual tokens. 
In this paper, we investigate layer-wise redundancy in MLLMs by introducing a novel metric, Layer Contribution (LC), which quantifies the impact of a layer's transformations on visual and text tokens, respectively.
The calculation of LC involves measuring the divergence in model output that results from removing the layer's transformations on the specified tokens.
Our pilot experiment reveals that many layers of MLLMs exhibit minimal contribution during the processing of visual tokens.
Motivated by this observation, we propose ShortV, a training-free method that leverages LC to identify ineffective layers, and freezes visual token updates in these layers.
Experiments show that ShortV can freeze visual token in approximately 60\% of the MLLM layers, thereby dramatically reducing computational costs related to updating visual tokens.
For example, it achieves a 50\% reduction in FLOPs on LLaVA-NeXT-13B while maintaining superior performance.
The code is publicly available at \href{https://github.com/icip-cas/ShortV}{https://github.com/icip-cas/ShortV}.

\end{abstract}
    
\section{Introduction}
\label{sec:intro}

Large language models (LLMs) have achieved remarkable performance in natural language tasks~\cite{yang2023baichuan2openlargescale, vicuna2023, openai2024gpt4technicalreport, llama3v, touvron2023llama2openfoundation}.
Building upon LLMs, Multimodal Large Language Models (MLLMs)~\cite{liu2024improvedbaselinesvisualinstruction, liu2024llavanext, gpt4v, 4oapi} take a significant step towards understanding the real physical world by incorporating visual information into their processes.
Typically, an MLLM consists of a visual encoder, a projector, and an LLM backbone.
Most of them preprocess visual information through a visual encoder, \eg a CLIP-ViT~\cite{radford2021learningtransferablevisualmodels, dosovitskiy2021imageworth16x16words}, and project the patch-level visual features into visual tokens through a projector.
Then they concatenate visual and text tokens and feed them into the LLM backbone.

\begin{figure}[t!]
\begin{subfigure}[b]{\linewidth}
    \centering
    \includegraphics[width=0.86\linewidth]{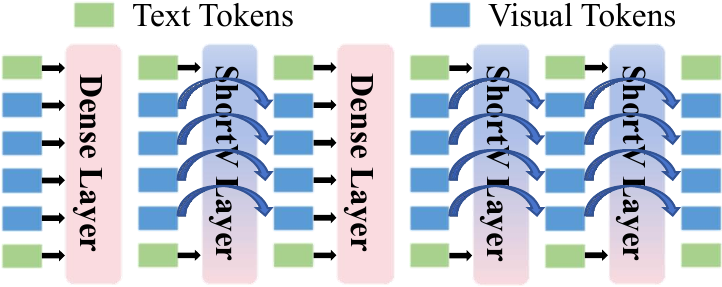}
    \caption{ShortV}
    \label{overview}
\end{subfigure}%
\hfill
\begin{subfigure}[b]{\linewidth}
    \centering
    \includegraphics[width=\linewidth]{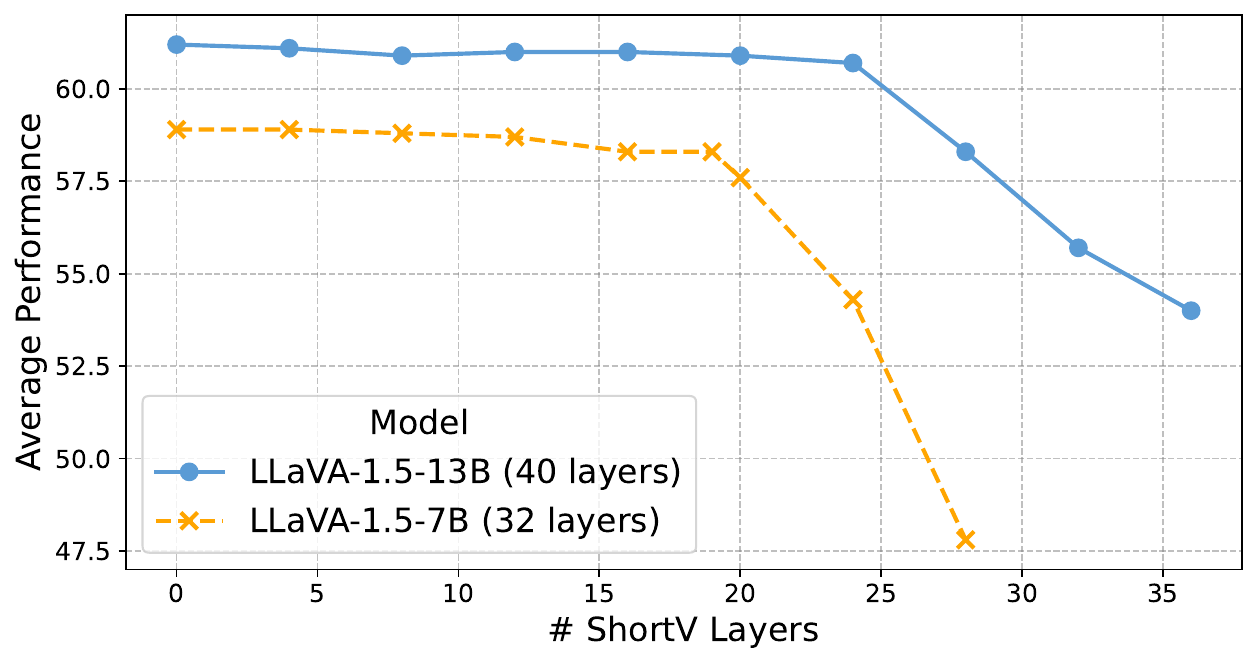}
    \caption{Performance \vs the Number of ShortV Layers}
    \label{vs}
\end{subfigure}%
\caption{
(a) \textbf{Illustration of ShortV.}
We identify ineffective layers for visual tokens and replace these layers with sparse ShortV layers.
In ShortV layers, we freeze visual tokens, and eliminate computations related to updating them.
ShortV improves MLLM efficiency in a training-free manner and involves no parameter updates.
Notably, ShortV is compatible with token pruning methods, \eg FastV.
(b) \textbf{Performance \vs the number of ShortV layers.} Average Performance means a normalized average score on multiple benchmarks.
ShortV can freeze visual tokens in approximately 60\% of the MLLM layers with nearly no performance degradation.
}
\label{fig:fig1}
\end{figure}

However, MLLMs face a substantial increase in computational overhead.
This burden primarily stems from the large scale of the LLM backbones and the significantly extended length of the concatenated visual-text token sequences.
To address this issue, Chen \etal~\cite{chen2024imageworth12tokens} discovers significant token-wise redundancy in MLLMs.
Based on this redundancy, they propose FastV, which identifies and prunes unimportant visual tokens in MLLMs to improve their efficiency.

Other than token-wise redundancy, in this paper, we reveal that MLLMs also exhibit significant layer-wise redundancy in processing visual tokens.
Specifically, we propose Layer Contribution (LC), a metric that quantifies how much a layer's transformations on certain tokens contribute to the model's output.
In the LC calculation of a layer, we freeze certain tokens in this layer, \ie keep the hidden states of the tokens unchanged, and then compute the Kullback-Leibler (KL) divergence between the resulting model's output logits and those of the original model.
This metric provides a direct measure of a layer's importance for certain tokens.
By comparing LC scores on visual tokens and those on text tokens, we discover that MLLM layers are ineffective for visual tokens, and their transformations on visual tokens contribute minimal to the model's output.

This phenomenon inspires us to propose ShortV, a simple but effective method to improve the efficiency of MLLMs.
In ShortV, we first utilize the LC metric to identify layers least effective at transforming visual tokens, and then replace these layers with sparse ShortV layers.
Within these sparse layers, visual tokens remain frozen, and the corresponding computations for updating them are eliminated, as shown in Figure~\ref{overview}.

To validate the effectiveness of ShortV, we conduct evaluations across multiple benchmarks, including MME~\cite{fu2024mmecomprehensiveevaluationbenchmark}, MMBench~\cite{liu2024mmbenchmultimodalmodelallaround}, MMMU~\cite{yue2024mmmumassivemultidisciplinemultimodal}, MMStar~\cite{chen2024rightwayevaluatinglarge}, SEED-Bench~\cite{li2023seedbenchbenchmarkingmultimodalllms}, GQA~\cite{hudson2019gqanewdatasetrealworld}, and Flickr30K~\cite{plummer2016flickr30kentitiescollectingregiontophrase}.
Figure~\ref{vs} illustrates the correlation between the normalized average performance on these benchmarks and the number of replaced layers.
As observed, ShortV can replace approximately 60\% of MLLM layers without performance degradation.
Unlike FastV and other token pruning methods, ShortV reduces computations of per visual token rather than reducing the number of visual tokens.
Therefore, ShortV and token pruning methods are orthogonal and compatible.
Furthermore, we demonstrate that combining ShortV and FastV can further enhance MLLM efficiency.

We summarize our contribution as follows.
\begin{itemize}
    \item We propose Layer Contribution (LC), a metric to quantify how much a layer's transformations on specific tokens contribute to the model's output.
    \item Leveraging LC, we reveal significant redundancy in MLLM layers for visual tokens. Transformations on visual tokens in many layers contribute minimally and are thus ineffective.
    \item Based on the observation above, we propose ShortV, which improves MLLM efficiency by freezeing visual tokens in ineffective layers. ShortV can freeze visual tokens in approximately 60\% of MLLM layers without performance degradation. Extensive experiments and ablation studies demonstrate ShortV’s effectiveness.
\end{itemize}

\section{Layer Redundancy in MLLMs}
\label{sec:motivation}

In this section, we first introduce the background of layer redundancy in text-only LLMs, and then propose a metric to measure MLLM layer redundancy for different types of tokens.
Next, we conduct a pilot experiment to investigate layer redundancy in MLLMs.

\subsection{Background}

Typically, MLLMs are built on text-only LLMs.
MLLMs employ pre-trained visual encoders, such as CLIP-ViT~\cite{radford2021learningtransferablevisualmodels, dosovitskiy2021imageworth16x16words}, to convert images into visual features, and then use projectors to project them into visual tokens in the text token embedding space.
The visual tokens are concatenated with text tokens and fed into the LLM backbones.

For text-only LLMs, Men \etal~\cite{men2024shortgptlayerslargelanguage} identify notable redundancy across their layer.
Some layers' transformations on the hidden states of text token contribute minimally to the overall model functionality.
Consequently, these layers are considered ineffective.
Removing these transformations in approximately 25\% of LLM layers has minimal impact on model outputs.
Such redundancy mainly occurs in middle-to-deeper layers, whereas initial layers and the last layer remain critical to the model functionality.
However, this pattern may not hold for MLLMs.
Huang \etal~\cite{huang2024decipheringcrossmodalalignmentlarge} demonstrate a clear modality gap in the embedding space of current MLLMs, where visual and text tokens exhibit a uniform distribution within each modality but a significant distribution gap between modalities.
Such a modality gap implies that MLLMs might adopt distinct computational patterns or strategies for processing visual and text tokens, potentially affecting how redundancy is distributed across layers.
This raises several key questions: Are MLLM layers as ineffective for visual tokens as LLM layers are for text tokens?
To what extent does layer redundancy exist in MLLMs?
How is this redundancy distributed across different MLLM layers?

\subsection{Layer Contribution Metric}
\label{lc}
\begin{figure}[t]
\centering
\begin{subfigure}[b]{\linewidth}
    \centering
    \includegraphics[width=\linewidth]{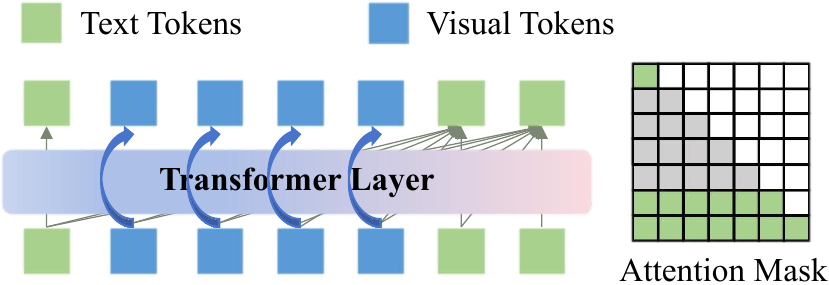}
    \caption{Sparse layer where \textbf{visual} tokens are frozen, used to investigate layer redundancy for \textbf{visual} tokens. 
    Only \textbf{text} tokens function as queries and are passed through the FFN.
    We also denote this layer as ShortV layer.}
    \label{visual}
\end{subfigure}%
\hfill
\begin{subfigure}[b]{\linewidth}
    \centering
    \includegraphics[width=0.985\linewidth]{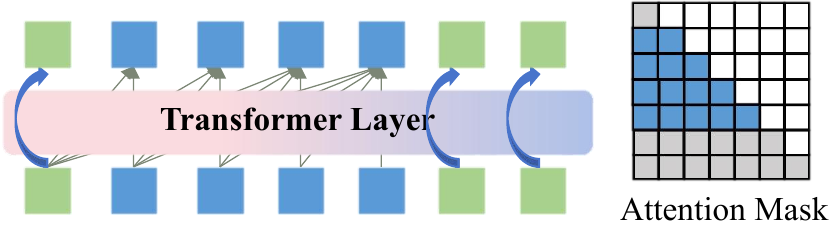}
    \caption{Sparse layer where \textbf{text} tokens are frozen, used to investigate layer redundancy for \textbf{text} tokens.
    Only \textbf{visual} tokens function as queries and are passed through the FFN.
    }
    \label{text}
\end{subfigure}%
\caption{Sparse layers used to investigate layer redundancy for different tokens. 
To investigate layer redundancy for certain tokens, we freeze these tokens within the layer, \ie keep hidden states of these tokens unchanged, and measure the divergence between the model's output logits and those of the original model.
We \textcolor{gray}{gray} out the attention that does not need calculation.
}
\label{fig:masks}
\end{figure}

To investigate layer redundancy for certain tokens, we freeze these tokens within the investigated layer, \ie keep hidden states of these tokens unchanged.
To achieve this, we introduce sparse layers shown in Figure~\ref{fig:masks} for visual and text tokens, respectively.
Based on these designs, we propose the Layer Contribution (LC) metric, which evaluates how much a layer's transformations on certain tokens contribute to the model's overall functionality.
In the calculation of LC, we replace the investigated dense layer with the corresponding sparse layer, and compute the Kullback-Leibler (KL) divergence between the resulting model's output logits and those of the original dense model.

Specifically, We assume the LLM backbone $M$ has $L$ layers, and each of them consists of‌ a self-attention block and a feed-forward network (FFN).
The input of the $i$-th layer at the $j$-th token position is $H_i^j$, and the corresponding output is $H_{i+1}^j$.
The output of the last layer, \ie the $L$-th layer, at the same position is $H_{L+1}^j$.
The model $M$ utilizes $H_{L+1}^{-1}$ at the last token position to compute logits for next-token prediction through the language model head $LM_{head}$:
\begin{equation}
logits(M)=LM_{head}(H_{L+1}^{-1}).
\end{equation}

To investigate how much the $i$-th layer's transformations on certain tokens $X$ contribute to the model functionality, we replace the $i$-th layer with a sparse layer where $X$ are frozen, \ie $X$'s hidden states remain unchanged in this sparse layer.
The resulting model is denoted as $\mathcal{M}_i^X$.
In practice, $X$ can be visual tokens $V$ or text tokens $T$.
As shown in Figure~\ref{fig:masks}, we introduce sparse layers where $V$ and $T$ are frozen, respectively.
In Figure~\ref{visual}, we freeze the visual tokens in the sparse layer.
Within the self-attention block of this layer, the visual tokens do not attend to other tokens, and only the text tokens function as queries.
For the FFN of this layer, we simply do not pass the visual tokens through it.
In Figure~\ref{text}, we freeze the text tokens in another sparse layer with similar designs.
Based on these, we define the $i$-th layer's Layer Contribution (LC) score for certain tokens $X$ as the KL divergence between the output logits of the original model $M$ and those of the model $\mathcal{M}_i^X$ where $X$ are frozen in the $i$-th layer:
\begin{equation}
LC_i^X = KL(logits(M), logits(\mathcal{M}_i^X)),
\end{equation}
here $KL(\cdot)$ denotes KL divergence.
A lower LC score implies that the layer's transformations on the tokens exhibit minimal contribution to the model's output, suggesting that these transformations are ineffective.

\vspace{-0.35cm}
\paragraph{Discussion: Why not use perplexity or cosine similarity as the metric to measure the importance of layers?}
Some work in text-only LLMs utilizes perplexity~\cite{song2024slebstreamliningllmsredundancy, kim2024shortenedllamadepthpruning} or cosine similarity~\cite{men2024shortgptlayerslargelanguage, chen2024streamlining, he2024matters} as metrics to measure the importance of each layer.
For the former metric, instead of KL divergence, they measure the change in perplexity of the the models, and the layers causing minimal perplexity changes are deemed ineffective.
This metric, however, is inadequate when measuring layer redundancy for visual tokens.
We find that even if we do not feed visual tokens into the MLLMs, they can still generate reasonable responses, and the changes in perplexity of the MLLMs is relatively low.
Nevertheless, they face significant performance degradation in vision language tasks when the visual information is absent.
Thus, perplexity is not a reliable measure when evaluating layer redundancy for visual tokens.

\begin{figure*}
  \centering
  \begin{subfigure}{0.49\linewidth}
    \includegraphics[height=4.5cm, width=\linewidth]{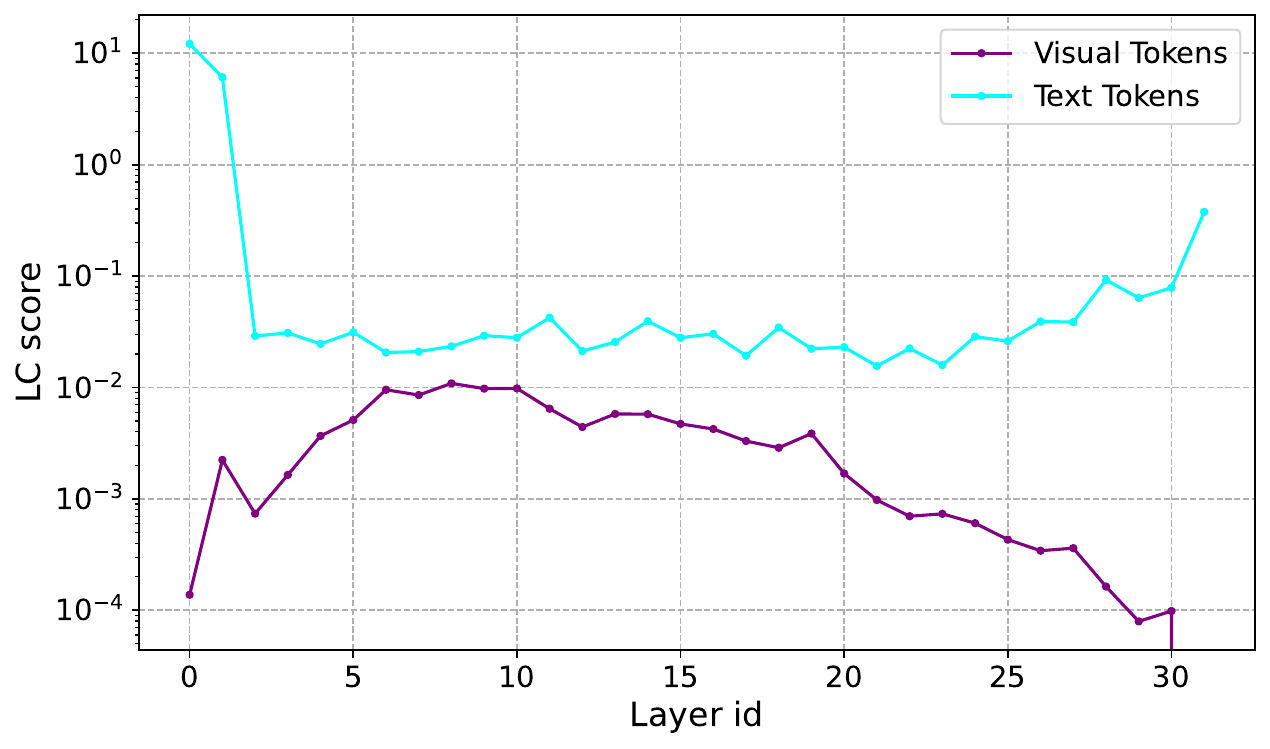}
    \caption{LLaVA-1.5-7B}
  \end{subfigure}
  \hfill
  \begin{subfigure}{0.49\linewidth}
    \includegraphics[height=4.5cm, width=\linewidth]{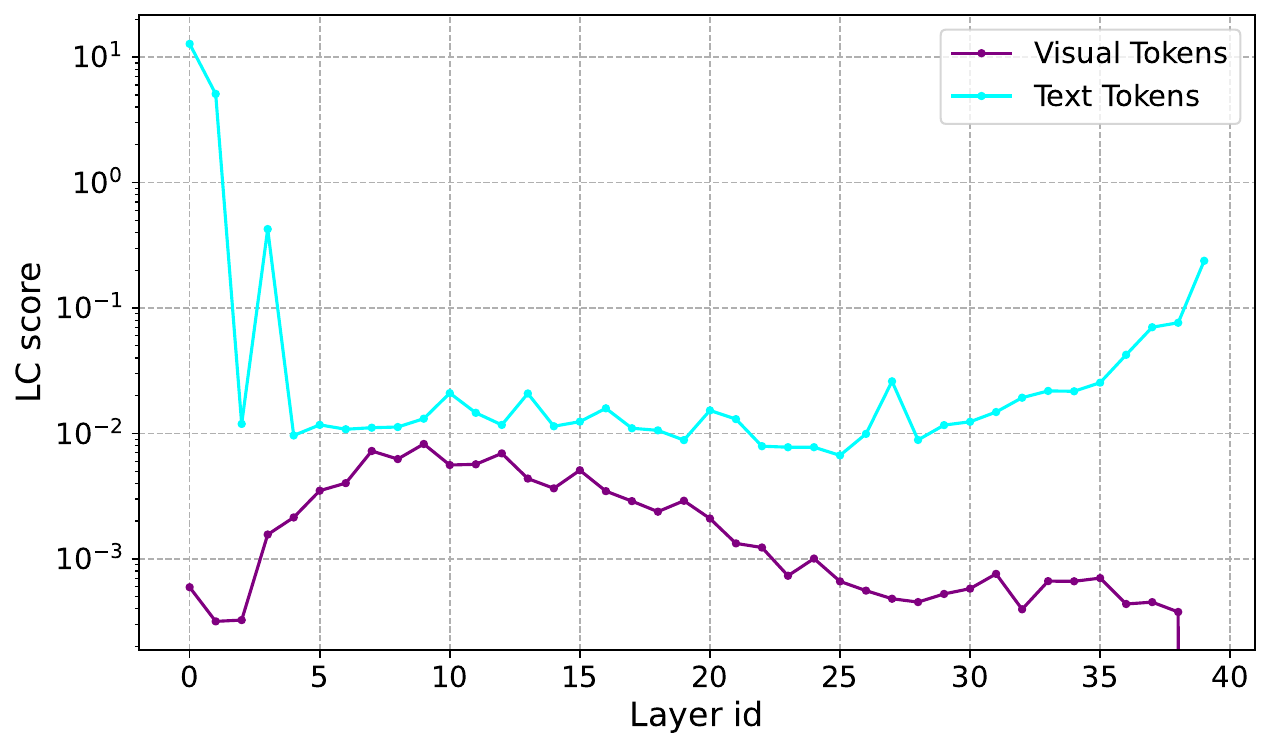}
    \caption{LLaVA-1.5-13B}
  \end{subfigure}
  \caption{The Layer Contribution (LC) scores of LLaVA-1.5-7B and LLaVA-1.5-13B.
  A lower LC score implies that the layer's transformations on the specified tokens are more ineffective.
  Layers are more ineffective for visual tokens than for text tokens, and freezing visual tokens in ineffective layers results in minimal output divergence from the original model.
  }
  \label{fig:lc}
\end{figure*}

For the latter metric, the cosine similarity between the input and output of a certain layer is calculated.
The hypothesis here is that the ineffective layers have less transformations on the hidden states of tokens, and therefore their inputs and outputs demonstrate higher similarities.
However, in our evaluation on LLaVA-1.5~\cite{liu2024improvedbaselinesvisualinstruction} models, the cosine similarity metric and LC differ in their measurement of the redundancy distribution across different layers.
Compared with the LC metric, which directly measures the logits divergence between model outputs, cosine similarity consistently overestimates the redundancy of the shallow layers and underestimates the redundancy of the deep layers.
We believe the reason behind this difference is that cosine similarity neglects the position of a layer within the model.
Specifically, minor transformations of hidden states in the shallow layers can influence all subsequent layers, whereas transformations with similar extent in deeper layers tend to have less impact on overall model functionality.

\subsection{Ineffective MLLM Layers for Visual Tokens}
\label{ineffective}

We conduct a pilot experiment to investigate layer redundancy in LLaVA-1.5-7B and LLaVA-1.5-13B~\cite{liu2024improvedbaselinesvisualinstruction}.
We first randomly sample 2,000 cases from a combination of two major vision language tasks, including caption (Flickr30K~\cite{plummer2016flickr30kentitiescollectingregiontophrase}) and visual question answering (GQA~\cite{hudson2019gqanewdatasetrealworld}).
Then we utilize these samples to calculate each MLLM layer's average LC score for visual and text tokens, respectively.
Figure~\ref{fig:lc} shows the results.
We summarize our findings as follows.

First, for text tokens, middle to deeper layers are more ineffective, while the initial and last layers make more contributions to the MLLM functionality. 
These observations align with the layer redundancy distribution of text-only LLMs found in Men \etal~\cite{men2024shortgptlayerslargelanguage}, indicating that visual instruction tuning~\cite{liu2023visualinstructiontuning, liu2024improvedbaselinesvisualinstruction} does not significantly alter the manner LLMs process text tokens.

Second, for visual tokens, the initial and the deep layers, including the last one, exhibit higher redundancy than other layers, which is different from the distribution for text token.
Notably, since the last layer's transformations on visual tokens do not contribute to the model's output, its LC score for visual tokens is always 0.

Third, layer redundancy shows an imbalance between visual and text tokens.
Each layer's LC score on visual tokens are lower than that on text tokens, which means that many layers' transformations on visual tokens are ineffective, and freezing visual token in these ineffective layers results in minimal impact on the models' output.

We attribute the different layer redundancy patterns for different modalities to the modality gap.
The clear distribution gap of visual and text tokens result in the difference in how MLLMs process them.
We hope these findings can provide insights into how MLLMs process visual and text tokens in different layers.

\section{ShortV}
\label{sec:method}

\begin{figure}[t]
  \centering
    \includegraphics[width=0.8\linewidth]{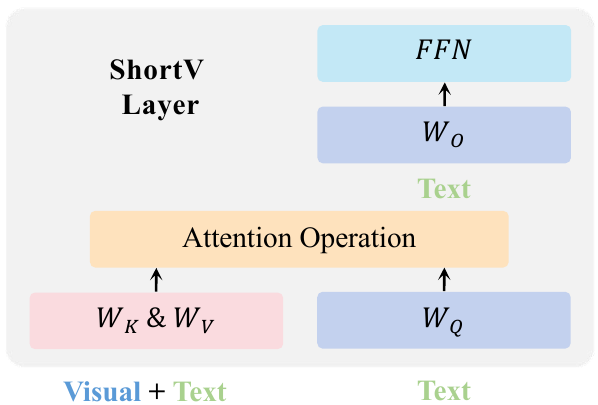}
   \caption{Details of ShortV layer. 
   In this layer, only text tokens pass through the $W_Q$ and $W_O$ matrices and the FFN.
   The attention mask is same as that in Figure~\ref{visual}, where visual tokens do not attend to other tokens, and only text tokens function as queries.
   }
   \label{fig:shortv}
\end{figure}

\begin{table*}[t]
  \centering
   \renewcommand{\arraystretch}{1.0}
  \resizebox{\linewidth}{!}
  {
  \begin{tabular}{lcccccccccc}
    \toprule
    \multirow{2}{*}{Method} & \multirow{2}{*}{TFLOPs} & FLOPs & \multirow{2}{*}{VQAv2} & \multirow{2}{*}{GQA}  & SEED- & MMMU & \multirow{2}{*}{MME} & MMBench  & \multirow{2}{*}{MMStar}   &  Flickr30K   \\ 
   & & Ratio & & & Bench & (val) & & EN  & & \textit{CIDEr}   \\
    \midrule
    \rowcolor{gray!15}\multicolumn{11}{c}{\textit{LLaVA-1.5-7B}} \\
    Vanilla & 8.5 & 100\% & 76.5 & 61.9   & 66.1 & 36.3 & 1510.7 & 64.1  & 33.7 & 74.9  \\ 
    \midrule
    FastV ($K$=2, $R$=50\%) & 4.9 & 58\% & 73.5  & 60.2 & 65.4 & 35.8 & 1475.6 & 64.3   & 32.4  & 67.5   \\
    VTW ($K$=16) & 4.7 & 55\% & 66.3 & 55.1 & \textbf{66.2} & 36.1 & 1497.0 & 64.0   & 32.8 & 44.5  \\
    ShortV (\textbf{Ours}, $N$=19) & 4.7 & 55\% & \textbf{75.7} & \textbf{60.9} & \textbf{66.2} & \textbf{36.2} & \textbf{1503.1} & \textbf{64.8}   & \textbf{33.3}  & \textbf{71.3}  \\
    \midrule
    \rowcolor{gray!15}\multicolumn{11}{c}{\textit{LLaVA-1.5-13B}} \\ 
     Vanilla & 16.6 & 100\% & 78.0  & 63.3 & 68.2 & 35.4 & 1531.3  & 68.9  & 36.1  & 79.6  \\
    \midrule
     FastV ($K$=2, $R$=50\%) & 9.4 & 57\% & 76.7 & 59.4  & 67.8 & 34.6 & 1506.6 & 68.3  & 35.9  & 73.4   \\
    VTW ($K$=20) & 9.1 & 55\% & 75.3 & 60.6  & \textbf{68.2} & 34.9 & 1533.0 & 68.5  & 36.1  & 65.9  \\
    ShortV (\textbf{Ours}, $N$=24) & 9.1 & 55\% & \textbf{77.2} & \textbf{62.0}  & 68.0 & \textbf{35.8} & \textbf{1535.9} & \textbf{68.6}  & \textbf{37.1} & \textbf{76.4}  \\ 
    \midrule
    \rowcolor{gray!15}\multicolumn{11}{c}{\textit{LLaVA-NeXT-7B}} \\ 
    
   Vanilla & 42.7 & 100\%  & 80.0  & 64.1  & 70.2 & 36.4 & 1519.0  & 67.1  & 37.1   & 69.7  \\
    \midrule
     FastV ($K$=2, $R$=50\%) & 22.0 & 52\% & \textbf{79.5} & 63.0  & 69.6 & 35.1 & 1482.0 & 66.3 & 36.5   & \textbf{66.7}   \\
      VTW ($K$=16) & 21.8 & 51\%  & 75.6 & 55.8  & 70.2 & 35.7 & 1518.2 & 67.1  & 37.6   & 38.4    \\
       ShortV (\textbf{Ours}, $N$=19) & 21.6 & 51\% & 78.8 & \textbf{63.4}  & \textbf{70.4} & \textbf{36.0} & \textbf{1525.1} & \textbf{67.2}  & \textbf{37.8}  & 65.7   \\
       \midrule
     \rowcolor{gray!15}\multicolumn{11}{c}{\textit{LLaVA-NeXT-13B}} \\  
     
     Vanilla & 81.8 & 100\%  & 80.9 & 65.7  & 71.9 & 35.9 & 1570.0 & 69.3  & 39.9   & 66.7 \\
    \midrule
      FastV ($K$=2, $R$=50\%) & 42.1 & 51\%  & 76.8 & 62.9  & 71.5 & 35.9 & 1546.4 & 68.5  & 39.6 & 66.0  \\
       VTW ($K$=20) & 41.7 & 51\% & 77.0 & 61.5  & \textbf{71.8} & 34.8 & \textbf{1569.4} & 69.1 & 39.8 & 56.6   \\
       ShortV (\textbf{Ours}, $N$=24) & 41.0 & 50\% & \textbf{79.7} & \textbf{63.6}  & \textbf{71.8} & \textbf{36.2} & 1553.0 & \textbf{70.2}  & \textbf{39.9}  & \textbf{67.5}  \\
    \bottomrule
  \end{tabular}%
  }
  \caption{Comparison of various training-free methods for MLLM efficiency. 
  FLOPs Ratio denotes the proportion of FLOPs retained after applying the corresponding method to improve MLLM efficiency, compared with the vanilla model.
  }
  \label{tab:main}
\end{table*}

\subsection{Freezing Visual Tokens in Ineffective Layers}
As demonstrated in the previous section, we identify significant layer redundancy for visual tokens in MLLMs.
Most layers' transformations on visual tokens are ineffective for the model functionality.
Based on this observation, we propose a direct method to enhance MLLM efficiency in a training-free manner: freezing visual tokens in ineffective layers.
We denote this method as \textbf{ShortV}.

In ShortV, we replace the ineffective dense layers with sparse ShortV layers, where visual tokens are frozen.
The ShortV layers are same as the sparse layer shown in ~\ref{visual},
and we illustrate their detailed architecture in Figure~\ref{fig:shortv}.
In each ShortV layer, only text tokens are passed through the FFN and the $W_Q$ and $W_O$ matrices of the attention block, and the attention mask is same as that in Figure~\ref{visual}, where visual tokens do not attend to other tokens, and only text tokens function as queries.

ShortV has one parameter: the number of replaced layers, which we denote as $N$.
First, we construct a tiny dataset, which contains a small number of samples from vision language tasks.
Then we use this dataset to calculate each layer's average LC score for visual tokens.
Next, we sort the layers in ascending order according to LC scores, and the layers with lower LC scores are more ineffective for visual tokens.
Finally, we freeze visual tokens in the $N$ layers with the lowest LC scores by replacing them with sparse ShortV layers, while keeping their original parameters.

ShortV is training-free and involves no parameter updates.
It can be applied to various MLLMs for different vision language tasks.
Notably, ShortV is orthogonal to and compatible with popular visual token pruning methods, \eg FastV~\cite{chen2024imageworth12tokens}.
Visual token pruning directly reduces the number of visual tokens, while ShortV mitigates the computational overhead related to each visual token.
This means that we can apply ShortV and token pruning at the same time to further improve MLLM efficiency.

\subsection{Computational Cost}
We consider the computations of self-attention blocks and feed-forward networks (FFNs) in layers of the LLM backbone.
Assume $t$ is the number of text tokens, $v$ is the number of visual tokens, $h$ is the hidden state size, $m$ is the intermediate size of the FFNs, the total FLOPs of one dense Transformer layer can be estimated as:
\begin{equation}
FLOPs=2(t+v)(4h+3m)h+4(t+v)^2h.
\end{equation}
For one ShortV layer, the FLOPs can be calculated as:
\begin{equation}
FLOPs^*=2t(4h+3m)h+4vh^2+4t(t+v)h.
\end{equation}
For the whole model, assume the LLM has $L$ layers in total, ShortV selects $N$ ineffective dense layers and replaces them with ShortV layers.
The FLOPs of the original dense model are $L\times FLOPs$, and the FLOPs of ShortV are calculated as $(L-N)\times FLOPs+N\times FLOPs^*$.
The FLOPs ratio of ShortV and the original model is computed as:
\begin{equation}
    r=\frac{(L-N)\times FLOPs+N\times FLOPs^*}{L\times FLOPs}.
\label{eq:ratio}
\end{equation}

\section{Experiments}
\label{sec:experiments}

\begin{table*}[t]
  \centering
  \small
   \renewcommand{\arraystretch}{1.0}
  \resizebox{0.9\linewidth}{!}
  {
  \begin{tabular}{cccccccccccc}
    \toprule
    \# ShortV  & \multirow{2}{*}{TFLOPs} & FLOPs & \multirow{2}{*}{MME} & MMBench & MMMU & \multirow{2}{*}{MMStar} & SEED-  & \multirow{2}{*}{GQA} & Flickr30K & \multirow{2}{*}{Avg.} & \multirow{2}{*}{Per.} \\ 
   Layers ($N$) &  & Ratio &  & EN & (val) &  & Bench  &  & \textit{CIDEr} \\
    \midrule
    \rowcolor{gray!15}\multicolumn{12}{c}{\textit{LLaVA-1.5-7B (32 layers)}} \\
    0 & 8.5 & 100\% & \textbf{1510.7} & 64.1 & \textbf{36.3} & 33.7 & 66.1  & \textbf{61.9} & \textbf{74.9} & 58.9 & 100.0 \\ 
    8 & 6.9 & 81\% & 1508.6 & 64.3 & 36.0 & \textbf{33.8} & \textbf{66.2} & 61.4 & 74.5 & 58.8 & 99.8 \\
    16 & 5.3 & 62\% & 1487.0 & \textbf{64.9} & 36.1 & 33.3 & 65.7  & 61.0 & 72.8 & 58.3 & 99.0 \\
    19 & 4.7 & 55\%  & 1503.1 & 64.8 & 36.2 & 33.3 & \textbf{66.2}  & 60.9 & 71.3 & 58.3 & 99.0 \\
    24 & 3.7 & 44\% & 1341.7 & 60.7 & 34.1 & 33.4 & 62.5  & 58.3 & 64.2 & 54.3 & 92.2 \\
    \midrule
    \rowcolor{gray!15}\multicolumn{12}{c}{\textit{LLaVA-1.5-13B (40 layers)}} \\ 
     0 & 16.6  & 100\% & 1531.3 & \textbf{68.9} & 35.4 & 36.1 & \textbf{68.2}  & \textbf{63.3} & \textbf{79.6} & 61.2 & 100.0 \\
   8 & 14.1 & 85\% & 1521.9 & 68.6 & 35.6 & 36.0 & \textbf{68.2} & 63.0 & 79.0 & 60.9 & 99.5 \\
    16 & 11.6 & 70\% & 1534.9 & 68.6 & \textbf{36.3} & 36.2 & 68.0  & 62.9  & 78.5 & 61.0 & 99.7 \\
    24 & 9.1 & 55\% & \textbf{1535.9} & 68.6 & 35.8 & \textbf{37.1} & 68.0  & 62.0 & 76.4 & 60.7 & 99.2 \\ 
    32 & 6.6 & 40\% & 1298.8 & 64.5 & 33.6 & 36.0 & 63.2 & 59.3 & 68.4 & 55.7 & 91.0 \\
    
    \midrule
    \rowcolor{gray!15}\multicolumn{12}{c}{\textit{LLaVA-NeXT-7B (32 layers)}} \\
    0 & 42.7 & 100\% & 1519.0 & 67.1 & 36.4 & 37.1 & 70.2  & \textbf{64.1} & 69.7 & 60.1 & 100.0 \\ 
    8 & 33.8 & 79\% & 1515.1 & \textbf{67.2} & \textbf{36.6} & 36.9 & 70.2 & \textbf{64.1} & \textbf{70.0} & 60.1 & 100.0 \\
    16 & 24.9 & 58\% & 1476.8 & \textbf{67.2} & 36.2 & 37.3 & 70.2  & 63.5 & 67.8 & 59.4 & 98.8 \\
    19 & 21.6 & 51\%  & \textbf{1525.1} & \textbf{67.2} & 36.0 & \textbf{37.8} & \textbf{70.4}  & 63.4 & 65.7 & 59.5 & 99.0 \\
    24 & 16.0 & 37\% & 1504.1 & 65.4 & 36.4 & 36.0 & 68.1 & 60.5 & 64.9 & 58.1 & 96.7 \\
    
    \midrule
    \rowcolor{gray!15}\multicolumn{12}{c}{\textit{LLaVA-NeXT-13B (40 layers)}} \\ 
     0 & 81.8  & 100\% & 1570.0 & 70.5 & 35.9 & 39.9 & \textbf{71.9}  & \textbf{65.7} & 66.7 & 61.3 & 100.0 \\
   8 & 68.2 & 83\% & 1552.4 & \textbf{70.6} & 35.0 & 39.6 & \textbf{71.9} & 65.1 & 66.9 & 61.0 & 99.5 \\
    16 & 54.6 & 67\% & 1561.0 & 70.1 & 35.0 & 39.7 & \textbf{71.9}  & 64.8  & 66.9 & 60.9 & 99.3 \\
    24 & 41.0 & 50\% & 1553.0 & 70.2 & \textbf{36.2} & 39.9 & 71.8  & 63.6 & 67.5 & 61.0 & 99.5 \\ 
    32 & 27.5 & 34\% & 1468.4 & 65.8 & 35.2 & 38.9 & 69.3 & 60.5 & 58.5 & 57.4 & 93.6 \\
    
    \bottomrule
  \end{tabular}%
  }
  \caption{Performance \vs Efficiency Balance of ShortV under different configurations.
\# ShortV Layers ($N$): the number of ShorV layers, Avg.: a normalized average score on the benchmarks, Per.: the relative performance retention compared with the vanilla models.
  }
  \label{tab:ratio}
\end{table*}

\subsection{Experimental Setups}

\paragraph{Models.} To validate the effectiveness of ShortV, we conduct experiments on popular open-source MLLMs, such as LLaVA-1.5-7B~\cite{liu2024improvedbaselinesvisualinstruction}, LLaVA-1.5-13B, LLaVA-NeXT-7B~\cite{liu2024llavanext} and LLaVA-NeXT-13B.
LLaVA-1.5 models process images with a 336$\times$336 resolution and treat each image as 576 tokens.
LLaVA-NeXT splits high-resolution images into subimages, and encode the subimages and down-sampled original images independently.
This allows the models to scale the input to any arbitrary resolution, without performing positional embedding interpolation for ViTs~\cite{dosovitskiy2021imageworth16x16words}.
LLaVA-NeXT scales the input image resolution to 4$\times$ and visual token number up to 5$\times$ compared with LLaVA-1.5, \ie 2880 tokens for each image.

\vspace{-0.35cm}
\paragraph{Baselines.}  
To evaluate the effectiveness of ShortV, which improves MLLM efficiency in a training-free manner, we compare it with popular training-free methods for MLLM efficiency, such as FastV~\cite{chen2024imageworth12tokens} and VTW~\cite{lin2025boostingmultimodallargelanguage}.
FastV drops visual tokens by a percentage of $R$ after the $K$-th layer in the forward process of input tokens.
It computes the average attention score one token received from all other tokens as the importance criterion to select pruned tokens.
VTW drops all visual tokens after the $K$-th layer, enabling only text tokens to engage in the subsequent layers.
We use the default settings for the baselines as in their original papers.
Specifically, for FastV, $K$=2 and $R$=50\%.
For VTW, $K$=16 for 7B models and $K$=20 for 13B models.

\subsection{Main Results}

In this section, we conduct experiments to compare ShortV with the baselines.
The results are shown in Table~\ref{tab:main}.
We perform evaluation on multiple popular vision language benchmarks, including MME~\cite{fu2024mmecomprehensiveevaluationbenchmark}, MMBench~\cite{liu2024mmbenchmultimodalmodelallaround}, MMMU (val)~\cite{yue2024mmmumassivemultidisciplinemultimodal}, MMStar~\cite{chen2024rightwayevaluatinglarge}, SEED-Bench~\cite{li2023seedbenchbenchmarkingmultimodalllms}, VQAv2~\cite{goyal2017makingvvqamatter}, and GQA~\cite{hudson2019gqanewdatasetrealworld}.
We manually choose the number of ShortV layers $N$ to maintain a similar or lower FLOPs ratio compared with the baselines.
The FLOPs ratio of ShortV is calculated according to Equation~\ref{eq:ratio}, where we set the number of text tokens to 64.
Specifically, we choose $N$=19 for the 7B models, and $N$=24 for the 13B models.
As shown in Table~\ref{tab:main}, our ShortV achieves comparable or superior performance across multiple benchmarks compared with the baselines.

\subsection{Balance between Efficiency and Performance}
In this section, we conduct an experiment to investigate the impact of ShortV's parameter $N$, which denotes the number of ShortV layers.
The experimental results are presented in Table~\ref{tab:ratio}. 
To facilitate intuitive comprehension, we plot the correlation between the normalized average score on benchmarks and the number of ShortV layers in Figure~\ref{vs}.

We observe that ShortV can freeze visual tokens in approximately 60\% of the MLLM layers while preserving superior performance.
As $N$ continues to increase, both 7B and 13B models can maintain more than 90\% performance when the hidden states of visual tokens remain unchanged in about 80\% of the layers.
These results are significantly different from those on text-only LLMs.
For LLMs, Men \etal~\cite{men2024shortgptlayerslargelanguage} remove transformations on text tokens in approximately 25\% of the LLM layers, and this results in about 10\% performance degradation on language benchmarks.
As the number of layers increases, the performance of LLMs rapidly declines.
These differences align with our observation in Section~\ref{ineffective} that layers are more ineffective for visual tokens than for text tokens.

In addition to the theoretical FLOPs ratios, we provide the speedups on real hardware using different settings, as shown in Table~\ref{tab:speedup} and Table~\ref{tab:speedup_13b}.
To get rid of the influence of different output sequence lengths, we use the first token latency to calculate the speedups.
We utilize the MMMU dataset for the latency test.
For comparison, we note that FastV~\cite{chen2024imageworth12tokens} with its default setting, \ie $K$=2 and $R$=50\%, achieves a 1.31$\times$ speedup over the vanilla LLaVA-1.5-13B model.
In contrast, our ShortV with its default parameter, \ie $N$=24, achieves a greater speedup of 1.39$\times$.

\begin{table}[t]
  \centering
    \resizebox{\linewidth}{!}
    {
    \begin{tabular}{lccccc}
    \toprule
    \# ShortV Layers ($N$) & 0 & 8 & 16 & 19 & 24 \\
    \midrule
    LLaVA-1.5-7B & 1.00$\times$ & 1.13$\times$ & 1.23$\times$ & 1.30$\times$ & 1.40$\times$  \\
    LLaVA-NeXT-7B & 1.00$\times$ & 1.15$\times$ & 1.35$\times$ & 1.44$\times$ & 1.64$\times$ \\
    \bottomrule
    \end{tabular}
    }
    \caption{Inference speedups over the vanilla models, based on the 7B models.
    We conduct this test on a single GPU.
    }
  \label{tab:speedup}
\end{table}

\begin{table}[t]
  \centering
    \resizebox{\linewidth}{!}
    {
    \begin{tabular}{lccccc}
    \toprule
    \# ShortV Layers ($N$) & 0 & 8 & 16 & 24 & 32 \\
    \midrule
    LLaVA-1.5-13B & 1.00$\times$ & 1.13$\times$ & 1.24$\times$ & 1.39$\times$ & 1.50$\times$  \\
    LLaVA-NeXT-13B & 1.00$\times$ & 1.13$\times$ & 1.30$\times$ & 1.52$\times$ & 1.84$\times$ \\
    \bottomrule
    \end{tabular}
    }
    \caption{Inference speedups over the vanilla models, based on the 13B models. We conduct this test on a single GPU.
    }
  \label{tab:speedup_13b}
\end{table}

\begin{table}[t]
  \centering
    \resizebox{\linewidth}{!}
    {
    \begin{tabular}{lccccc}
    \toprule
    \multirow{2}[2]{*}{Method} & FLOP & MMBench & MMMU  & SEED- & \multirow{2}[2]{*}{GQA}  \\
      & Ratio  & EN & (val)  & Bench &                 \\
    \midrule
    Vanilla & 100\% & 64.0 & 36.3 & 66.1 & 61.9  \\
    \midrule
    FastV & 58\% & 64.3 & 35.8 & 65.4 & 60.2  \\
    ShortV & 55\% & 64.8 & 36.2 & 66.2 & 60.9 \\
    ShortV+FastV & 29\% & 64.2 & 37.1 & 65.1 & 59.3 \\
    \bottomrule
    \end{tabular}
    }
    \caption{ShortV is compatible with FastV, and applying both at the same time can further enhance MLLM efficiency.
    This experiment is based on LLaVA-1.5-7B.
    }
  \label{tab:compatible}
\end{table}

\subsection{Orthogonal to Token Pruning}
In this section, we demonstrate that ShortV is orthogonal to and compatible with visual token pruning, \eg FastV~\cite{chen2024imageworth12tokens}.
FastV identifies $R$\% unimportant visual tokens and drops them after the $K$-th layer in the forward process of input tokens.
We apply FastV to ShortV, which already replaces $N$ ineffective layers for visual tokens with ShortV layers.
We use the default settings for FastV and ShortV in this experiment, \ie $K$=2 and $R$=50\% for FastV, and $N$=19 for ShortV.
We employ LLaVA-1.5-7B as the vanilla model.
The experimental results in Table~\ref{tab:compatible} demonstrate that ShortV is compatible with FastV and that the application of both can further improve MLLM efficiency.

\begin{table}[t]
  \centering
    \resizebox{\linewidth}{!}
    {
    \begin{tabular}{lccccc}
    \toprule
    \multirow{2}[2]{*}{Strategy} & FLOP & MMBench & MMMU  & SEED- & \multirow{2}[2]{*}{GQA}  \\
      & Ratio  & EN & (val)  & Bench &                 \\
    \midrule
    Vanilla & 100\% & 64.0 & 36.3 & 66.1 & 61.9  \\
    \midrule
    Random & 55\% & 58.4 & 33.6 & 60.5 & 56.1 \\
    Cosine Sim. & 55\% & 60.8 & 34.2 & 62.7 & 59.5 \\
    LC (Ours) & 55\% & 64.8 & 36.2 & 66.2 & 60.9 \\
    \bottomrule
    \end{tabular}
    }
    \caption{Ablation on strategies to select replaced layers, based on LLaVA-1.5-7B.
    ``Random" denotes randomly selecting 19 layers and freezing visual tokens in them.
    ``Cosine Sim." denotes using cosine similarity to select ineffective layers for visual tokens.
    }
  \label{tab:metric}
\end{table}

\subsection{Ablation Studies}
\paragraph{Ablation on strategies to select replaced layers.}
In this paragraph, we perform an ablation experiment on LLaVA-1.5-7B to investigate the impact of strategies for selecting which layers to replace.
ShortV selects ineffective layers for visual tokens, and replace them with ShortV layers.
To identify which layers are ineffective, we utilize the LC metric introduced in Section~\ref{sec:motivation}.
In contrast, previous work~\cite{men2024shortgptlayerslargelanguage} on text-only LLMs uses a metric based on cosine similarity.
It calculates the average cosine similarity between the inputs and outputs of each layer.
The layers with higher cosine similarities are deemed more ineffective.
To make a comparison between this cosine similarity metric and our LC metric, we calculate each layer's cosine similarity between the input hidden states and output hidden states of visual tokens, and select the same number of layers with the highest cosine similarities.
We show the comparison in Table~\ref{tab:metric}.
We also include the results of another baseline, ShortV (Random), where visual tokens are frozen in the same number of randomly selected layers.
These results clearly demonstrate that our LC metric performs better than cosine similarity in identifying ineffective MLLM layers for visual tokens, and ShortV based on the LC metric achieves performance comparable to the vanilla model.
In contrast, ShortV based on the cosine similarity metric cannot match the performance of the vanilla model, although it outperforms the baseline with randomly selected layers.

\vspace{-3.5mm}
\paragraph{Ablation on frozen tokens.}
In this paragraph, we conduct an ablation study on LLaVA-1.5-7B to investigate the impact of freezing different types of tokens.
In Section~\ref{sec:motivation}, we demonstrate that MLLM layers are ineffective for visual tokens, as measured by the LC metric.
Motivated by this observation, ShortV freezes visual tokens in ineffective layers.
In Table~\ref{tab:token}, we compare our method with the strategies of freezing other tokens.
In the experiment detailed in line (a), we utilize the LC metric to identify 19 ineffective layers for text tokens, and freeze text tokens in them.
Despite having fewer frozen tokens, we can observe that this strategy results in significant performance declines compared with our method, which freezes visual tokens rather than text tokens.
These experimental results align with our findings in Section~\ref{sec:motivation} that MLLM layers are more ineffective for visual tokens than for text tokens.
In lines (b) and (c), we first calculate each layer's average LC score for all tokens, including visual and text tokens, and then select 19 ineffective layers.
Next, in the experiment corresponding to line (b), we freeze all input tokens in these layers. 
In line (c), we freeze random input tokens, and the number of frozen tokens matches that of the visual tokens. 
As a result, the computational overhead associated with line (c) is the same as that of our method in line (d).
We can find that freezing tokens other than visual tokens leads to substantial performance degradation in vision-language tasks.
These ablations demonstrate the effectiveness of our ShortV.

\begin{table}[t]
  \centering
    \scalebox{0.8}
    {
    \begin{tabular}{lcccc}
    \toprule
    \multirow{2}[2]{*}{Frozen Tokens}  & MMBench & MMMU  & SEED- & \multirow{2}[2]{*}{GQA}  \\
      & EN & (val)  & Bench &                 \\
    \midrule
    None (Vanilla)  & 64.0 & 36.3 & 66.1 & 61.9  \\
    \midrule
    (a) Text  & 2.1 & 23.7 & 8.9 & 2.9 \\
    (b) Text+Visual  & 1.3 & 26.6 & 0.8 & 0.0 \\
    (c) Random  & 1.5 & 22.9 & 5.5 & 2.3 \\
    (d) Visual (Ours)  & 64.8 & 36.2 & 66.2 & 60.9 \\
    \bottomrule
    \end{tabular}
    }
    \caption{Ablation on frozen tokens, based on LLaVA-1.5-7B. (a) identifying 19 ineffective layers for text tokens and freezing text tokens in them.
    In lines (b) and (c), we select ineffective layers for all tokens.
    line (b) involves freezing all input tokens in them, whereas line (c) denotes randomly freezing the same number of tokens as the visual tokens.
    }
  \label{tab:token}
\end{table}

\section{Related Work}

\subsection{Multimodal Large Language Models}
Built upon Large Language Models (LLMs)~\cite{yang2023baichuan2openlargescale, vicuna2023, openai2024gpt4technicalreport, llama3v, touvron2023llama2openfoundation}, Multimodal Large Language Models (MLLMs)~\cite{fuyu-8b, lu2024deepseekvlrealworldvisionlanguageunderstanding, chen2023minigptv2largelanguagemodel, zhu2023minigpt4enhancingvisionlanguageunderstanding, ye2023mplugowl2revolutionizingmultimodallarge, wang2024cogvlmvisualexpertpretrained} have made significant progress in processing and understanding the visual world.
Typically, they use a decoder-only architecture.
Specifically, they utilize visual encoders~\cite{dosovitskiy2021imageworth16x16words, zhai2023sigmoidlosslanguageimage, radford2021learningtransferablevisualmodels, liu2022convnet2020s, ilharco2021openclip} to convert input visual information into visual features and then use projectors to project these visual features into visual tokens. 
These visual tokens are then concatenated with text tokens and fed into the LLM backbones.
Current MLLMs use hundreds to thousands of visual tokens to represent a single image, significantly increasing the length of the token sequences.
For instance, the LLaVA-1.5 models~\cite{liu2024improvedbaselinesvisualinstruction} transform each image with 336$\times$336 resolution into 576 tokens.
For images with higher resolutions, the LLaVA-NeXT models~\cite{liu2024llavanext} process images into up to 2,880 visual tokens, and SPHINX-2k~\cite{lin2023sphinxjointmixingweights} divides one image into nine subimages, resulting in 2,890 visual tokens.
Applying LLMs to such large numbers of visual tokens incurs substantial computational costs.
In this paper, we introduce ShortV to enhance the efficiency of MLLMs by reducing the computational overhead associated with visual tokens.

\subsection{Efficient LLMs and MLLMs}
For LLMs, previous studies~\cite{men2024shortgptlayerslargelanguage, song2024slebstreamliningllmsredundancy} find that layers in LLMs are ineffective for text tokens.
They remove computations in about 25\% of the layers, while preserving approximately 90\% of the performance.
LaCo~\cite{yang2024lacolargelanguagemodel} utilizes layer merging for efficient LLMs.

To address the computational inefficiency of MLLMs, previous methods~\cite{ren2025vamba, yuan2025saisamultimodallargelanguage, song2024moviechat, song2024moviechat+, chen2024imageworth12tokens, lin2025boostingmultimodallargelanguage} have primarily focused on two aspects: efficient model architecture and visual token compression.
Among efficient model architectures, cross-attention-based models~\cite{chen2024evlmefficientvisionlanguagemodel, alayrac2022flamingovisuallanguagemodel, awadalla2023openflamingoopensourceframeworktraining, idefics} insert gated cross-attention layers within LLM layers for visual perception, but previous studies~\cite{dai2024nvlmopenfrontierclassmultimodal, laurençon2024mattersbuildingvisionlanguagemodels} demonstrate that this architecture performs worse than the decoder-only architecture in the same settings.
Instead of inserting cross-attention layers, mPLUG-Owl3~\cite{ye2024mplugowl3longimagesequenceunderstanding} and Vamba~\cite{ren2025vamba} introduce cross-attention operations in parallel with self-attention.
In contrast, SAISA~\cite{yuan2025saisamultimodallargelanguage} introduces NAAViT self-attention blocks, which incorporate multimodal cross-attention into the original self-attention operations of the LLMs, and reuse the parameters of self-attention blocks.
The design of ShortV layers is inspired by NAAViT.
Differently, in ShortV layers, visual tokens also skip their FFNs.

Visual token compression methods improve MLLM efficiency in both training-based~\cite{chai2025auroracapefficientperformantvideo, li2023llamavidimageworth2, li2024tokenpackerefficientvisualprojector, jaegle2021perceivergeneralperceptioniterative, cha2024honeybeelocalityenhancedprojectormultimodal} and training-free~\cite{ye2024vocollamavisioncompressionlarge, shang2024llavaprumergeadaptivetokenreduction, chen2024imageworth12tokens} manners.
FastV~\cite{chen2024imageworth12tokens} reveals token-wise redundancy, and it removes unimportant tokens during inference.
In this paper, we reveal layer-wise redundancy in MLLMs.
Layers in MLLMs are much more ineffective for visual tokens than for text token.
Therefore, we can freeze visual tokens in approximately 60\% of the MLLM layers with minimal performance degradation.
Unlike previous methods for MLLM efficiency, ShortV does not reduce the number of visual tokens but instead decreases the computational costs of processing each token.
ShortV is training-free and orthogonal to token compression.
We demonstrate that ShortV is compatible with FastV, allowing for simultaneous application to further enhance MLLM efficiency.

\section{Conclusion}

In this paper, we explore the layer-wise redundancy in MLLMs.
We discover that layers in MLLMs are more ineffective for visual tokens than for text tokens.
MLLM layers' transformations on visual tokens have a minimal impact on the MLLM output.
Motivated by this observation, we propose ShortV, a training-free method to enhance MLLM efficiency.
ShortV utilizes our proposed LC metric to select ineffective layers for visual tokens, and freezes visual tokens in these layers.
It can freeze visual tokens in about 60\% of the layers while preserving superior performance.

\section{Acknowledgment}
We sincerely thank the reviewers for their insightful comments and valuable suggestions. This work was supported by Beijing Natural Science Foundation (L243006), Beijing Municipal Science and Technology Project (Nos. Z231100010323002), the Natural Science Foundation of China (No.  62306303, 62476265) and the Basic Research Program of ISCAS (Grant No. ISCAS-ZD-202401).

{
    \small
    \bibliographystyle{ieeenat_fullname}
    \bibliography{main}
}
\clearpage

\appendix

\section{Limitations and Future Work}
Despite the effectiveness of ShortV, It remains a coarse-grained method, and there are several directions to improve it.
First, ShortV treats each layer as a whole, whereas LLM layers have a more fine-grained structure, including attention blocks and FFNs, and He \etal~\cite{he2024matters} reveal that they exhibit different levels of redundancy in text-only LLMs.
Freezing visual tokens in different proportions of attention blocks and FFNs could achieve a more favorable balance between performance and efficiency.
Second, Chen~\etal~\cite{chen2024streamlining} uses a small network to update tokens in ineffective layers of LLMs, which is also a promising path to improve the performance of ShortV.

\section{Replaced Layers}
\label{replaced}

For the LC metric calculation to select the replaced layers, we randomly sample 40 cases from GQA and Flickr30K, with 20 from each of them.
Layers are replaced with ShortV layers in ascending order based on their LC values, starting from the lowest and moving to the highest.
In Table~\ref{layers}, we list the layer ids of replaced layers in default ShortV.

\begin{table}[h]
			\centering
			\small
			\begin{tabular}{ll} %
				\toprule
				Model & Replaced Layers \\
				\midrule
                LLaVA-1.5-7B & 31, 29, 30, 28, 0, 26, 27, 25, 24, 22,\\
                & 23, 21, 2, 3, 20, 18, 17, 12, 19 \\
                \midrule
                LLaVA-1.5-13B & 39, 32, 28, 36, 27, 37, 29, 30, 1, 38, \\
                & 25, 31, 2, 26, 23, 34, 0, 33, 35, 22, \\
                & 24, 21, 20, 17 \\
                \midrule
                LLaVA-NeXT-7B & 31, 29, 30, 28, 26, 27, 22, 24, 21, 23, \\
                & 25, 20, 19, 17, 18, 15, 12, 0, 2 \\
                \midrule
                LLaVA-NeXT-13B & 39, 32, 29, 36, 27, 30, 37, 23, 25, 31, \\
                & 26, 2, 28, 22, 33, 35, 34, 24, 38, 21, \\
                & 20, 18, 1, 17 \\
				\bottomrule
			\end{tabular}
            \caption{Replaced layers for different MLLMs.}
			\label{layers}
		\end{table}

\begin{table}[t]
  \begin{minipage}{0.99\linewidth}
\centering
\resizebox{\textwidth}{!}{%
\begin{tabular}{p{1.65cm} p{7.65cm} }
\toprule
&  \includegraphics[height=5cm]{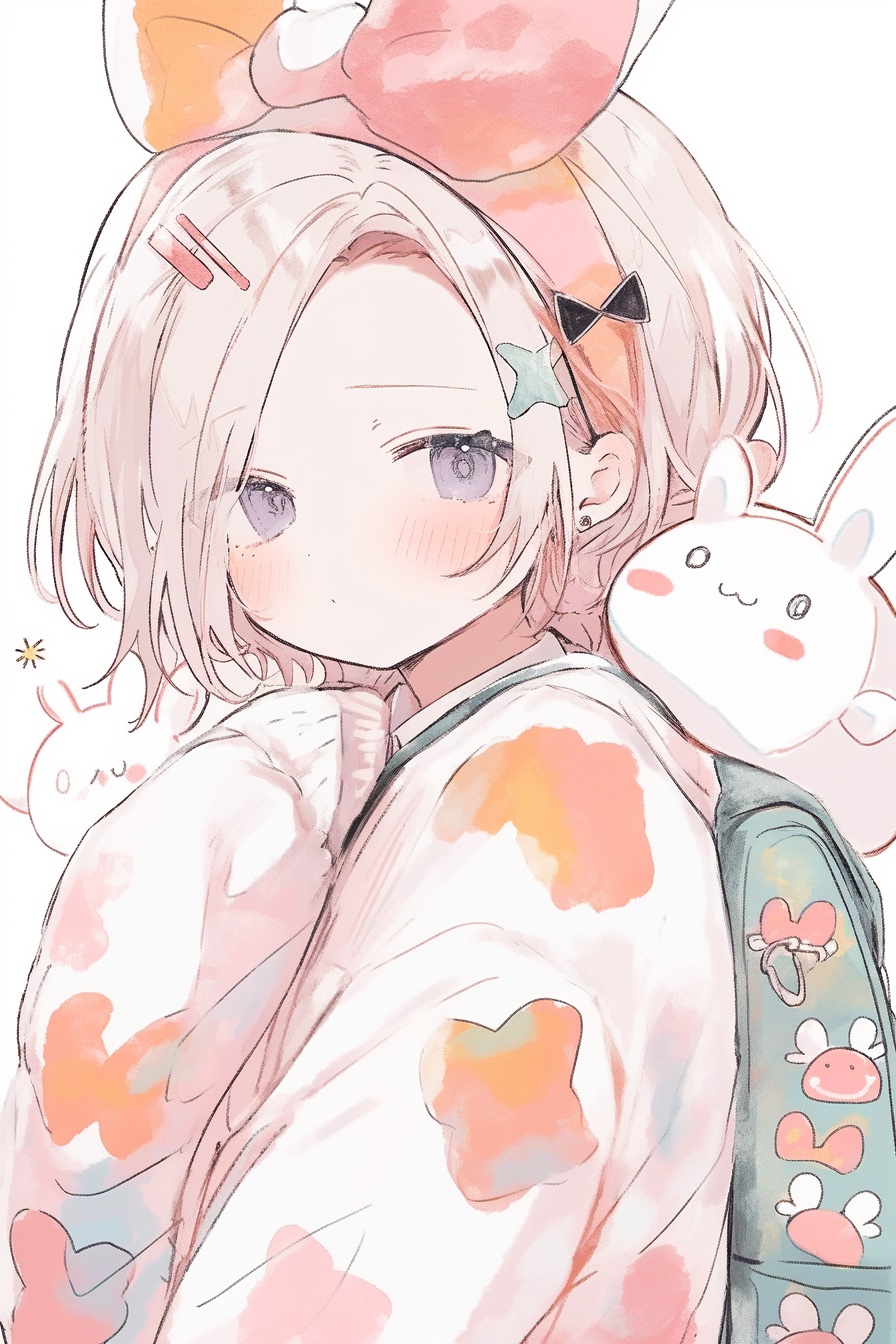} \\

User & Here's a formula for a Stable Diffusion image prompt: a cartoon-style image of {\color{orange}[adjective]} {\color{blue}[subject]} {\color{purple}[haircut]} {\color{brown}[clothing]} {\color{red}[doing action]}, {\color{green}[color scheme]}, detailed, trending on artstation. Generate the prompt for this image. \\
\midrule
w/o ShortV & A cartoon-style image of {\color{orange}a young} {\color{blue}girl} {\color{purple}with a bob haircut}, {\color{brown}wearing a white kimono}, {\color{red}holding a stuffed animal, and posing for a picture}, {\color{emerald}with a color scheme of pink and white}, detailed, and trending on artstation. \\
\midrule
w/ ShortV & A cartoon-style image of {\color{orange}a young} {\color{blue}girl} {\color{purple}with a bob haircut}, {\color{brown}wearing a pink and white kimono}, {\color{red}holding a stuffed animal, and posing for a picture}, {\color{emerald}with a color scheme of pink and white}, detailed, and trending on artstation. \\
\bottomrule
\end{tabular}
}
\captionof{table}{Responses comparison between the vanilla LLaVA-1.5-13B model and the LLaVA-1.5-13B model with ShortV ($N$=24), on constrained Stable Diffusion prompt generation.
}
\label{tab:anime_prompt_generation}
  \end{minipage}
\end{table}

\begin{table}
  \begin{minipage}{0.99\linewidth}
\centering
\resizebox{\textwidth}{!}{%
\begin{tabular}{p{1.65cm} p{7.65cm} }
\toprule
&  \includegraphics[height=3.5cm]{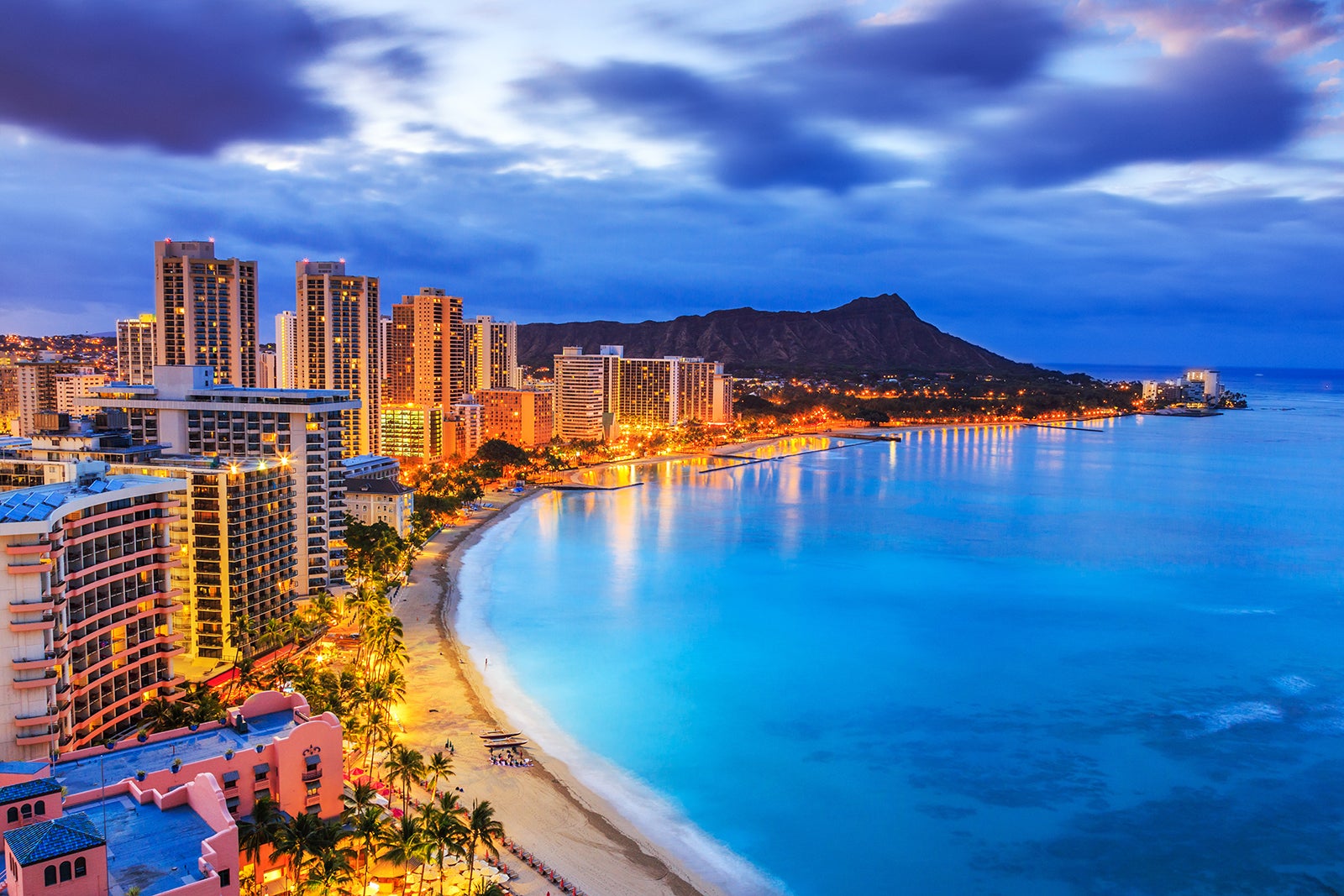} \\

User & If there are factual errors in the questions, point it out; if not, proceed to answering the question. \\
& What's happening in the desert? \\
\midrule
w/o ShortV & There are no deserts in the image. The image features  a beach with palm trees, a city skyline, and a large body of water. \\
\midrule
w/ ShortV & There are no deserts in the image. The image features a beach with a large body of water, a city skyline, and palm trees. \\
\bottomrule
\end{tabular}
}
\captionof{table}{Responses comparison between the vanilla LLaVA-1.5-13B model and the LLaVA-1.5-13B model with ShortV ($N$=24), on a tricky question.}
\label{tab:tricky_example}  
  \end{minipage}
\end{table}

\section{Qualitative Results}
In this section, we use the qualitative examples in the original LLaVA-1.5 paper to compare the responses of LLaVA-1.5-13B before and after applying ShortV.
We reference the responses of the vanilla model from the original LLaVA-1.5 paper.
We utilize the default parameter, \ie $N$=24, for the model with ShortV.
In Table~\ref{tab:anime_prompt_generation} and \ref{tab:tricky_example}, we can see that the model with ShortV generates responses similar to the vanilla model, despite freezing visual tokens in 24 MLLM layers.
These examples validate the capability of ShortV in enhancing MLLM efficiency while preserving performance.

\section{ShortV under Different Settings}
\label{settings}
\begin{table*}[t]
  \centering
  \small
   \renewcommand{\arraystretch}{1.0}
  \scalebox{1.0}
  {
  \begin{tabular}{cccccccccccc}
    \toprule
    \# ShortV  & \multirow{2}{*}{TFLOPs} & FLOPs & \multirow{2}{*}{MME} & MMBench & MMMU & \multirow{2}{*}{MMStar} & SEED-  & \multirow{2}{*}{GQA} & Flickr30K & \multirow{2}{*}{Avg.} & \multirow{2}{*}{Per.} \\ 
   Layers ($N$) &  & Ratio &  & EN & (val) &  & Bench  &  & \textit{CIDEr} \\
    \midrule

    \rowcolor{gray!15}\multicolumn{12}{c}{\textit{LLaVA-1.5-7B (32 layers)}} \\
    0 & 8.5 & 100\% & \textbf{1510.7} & 64.1 & \textbf{36.3} & 33.7 & 66.1  & \textbf{61.9} & \textbf{74.9} & 58.9 & 100.0 \\ 
    4 & 7.7 & 91\% & 1507.5 & 64.1 & \textbf{36.6} & 33.5 & \textbf{66.2}  & \textbf{61.9} & 74.7 & 58.9 & 100.0 \\
    8 & 6.9 & 81\% & 1508.6 & 64.3 & 36.0 & 33.8 & \textbf{66.2} & 61.4 & 74.5 & 58.8 & 99.8 \\
    12 & 6.1 & 72\% & 1495.2 & 64.2 & 36.2 & 34.0 & \textbf{66.2}  & 61.2 & 74.1 & 58.7 & 99.7 \\
    16 & 5.3 & 62\% & 1487.0 & \textbf{64.9} & 36.1 & 33.3 & 65.7  & 61.0 & 72.8 & 58.3 & 99.0 \\
    19 & 4.7 & 55\%  & 1503.1 & 64.8 & 36.2 & 33.3 & \textbf{66.2}  & 60.9 & 71.3 & 58.3 & 99.0 \\
    20 & 4.5 & 53\% & 1466.8 & 63.4 & 35.3 & \textbf{34.7} & 65.2  & 60.4 & 70.7 & 57.6 & 97.8 \\
    24 & 3.7 & 44\% & 1341.7 & 60.7 & 34.1 & 33.4 & 62.5  & 58.3 & 64.2 & 54.3 & 92.2 \\
    28 & 2.9 & 34\% & 1079.0 & 57.9 & 31.0 & 30.2 & 56.0  & 52.0 & 53.6 & 47.8 & 81.2 \\
    \midrule
    \rowcolor{gray!15}\multicolumn{12}{c}{\textit{LLaVA-1.5-13B (40 layers)}} \\ 
     0 & 16.6  & 100\% & 1531.3 & \textbf{68.9} & 35.4 & 36.1 & \textbf{68.2}  & \textbf{63.3} & \textbf{79.6} & 61.2 & 100.0 \\
     4 & 15.3 & 92\% & 1521.6 & 68.6 & 35.8 & 36.5 & \textbf{68.2} & \textbf{63.3} & 79.4 & 61.1 & 99.8 \\
   8 & 14.1 & 85\% & 1521.9 & 68.6 & 35.6 & 36.0 & \textbf{68.2} & 63.0 & 79.0 & 60.9 & 99.5 \\
    12 & 12.8 & 77\% & 1521.9 & 68.6 & 35.9 & 36.2 & 68.1  & 62.9 & 78.9 & 61.0 & 99.7 \\ 
    16 & 11.6 & 70\% & 1534.9 & 68.6 & \textbf{36.3} & 36.2 & 68.0  & 62.9  & 78.5 & 61.0 & 99.7 \\
    20 & 10.3 & 62\% & 1533.0 & 68.6 & 36.1 & 36.8 & 68.0  & 62.4 & 77.5 & 60.9 & 99.5 \\
    24 & 9.1 & 55\% & \textbf{1535.9} & 68.6 & 35.8 & \textbf{37.1} & 68.0  & 62.0 & 76.4 & 60.7 & 99.2 \\ 
    28 & 7.8 & 47\% & 1417.6 & 65.5 & 34.6 & 35.9 & 65.4  & 60.8 & 74.9 & 58.3 & 95.3 \\
    32 & 6.6 & 40\% & 1298.8 & 64.5 & 33.6 & 36.0 & 63.2 & 59.3 & 68.4 & 55.7 & 91.0 \\
    36 & 5.3 & 32\% & 1259.6 & 62.9 & 33.2 & 34.9 & 62.5  & 58.7 & 62.8 & 54.0 & 88.2 \\

    \midrule
    \rowcolor{gray!15}\multicolumn{12}{c}{\textit{LLaVA-NeXT-7B (32 layers)}} \\
    0 & 42.7 & 100\% & 1519.0 & 67.1 & 36.4 & 37.1 & 70.2  & \textbf{64.1} & 69.7 & 60.1 & 100.0 \\ 
    4 & 38.3 & 90\% & 1519.3 & \textbf{67.2} & \textbf{36.8} & 36.8 & \textbf{70.7} & \textbf{64.1} & 69.3 & 60.1 & 100.0 \\
    8 & 33.8 & 79\% & 1515.1 & \textbf{67.2} & 36.6 & 36.9 & 70.2 & \textbf{64.1} & 70.0 & 60.1 & 100.0 \\
    12 & 29.4 & 69\% & 1476.8 &  67.1 & 36.6 & 37.4 & 70.2 & 63.4 & \textbf{70.3} &  59.8 & 99.5 \\
    16 & 24.9 & 58\% & 1476.8 & \textbf{67.2} & 36.2 & 37.3 & 70.2  & 63.5 & 67.8 & 59.4 & 98.8 \\
    19 & 21.6 & 51\%  & \textbf{1525.1} & \textbf{67.2} & 36.0 & \textbf{37.8} & 70.4  & 63.4 & 65.7 & 59.5 & 99.0 \\
    20 & 20.5 & 48\% & 1505.6 & 66.7 & 36.3 & 37.3 & 70.0 & 63.0 & 65.5 & 59.2 & 98.5 \\
    24 & 16.0 & 37\% & 1504.1 & 65.4 & 36.4 & 36.0 & 68.1 & 60.5 & 64.9 & 58.1 & 96.7 \\
    \midrule
    \rowcolor{gray!15}\multicolumn{12}{c}{\textit{LLaVA-NeXT-13B (40 layers)}} \\ 
     0 & 81.8  & 100\% & 1570.0 & 70.5 & 35.9 & 39.9 & \textbf{71.9}  & \textbf{65.7} & 66.7 & 61.3 & 100.0 \\
     4 & 75.0 & 92\% & \textbf{1574.8} & \textbf{70.6} & 34.8 & 39.7 & \textbf{71.9} & 65.4 & 66.5 & 61.1 & 99.7 \\
   8 & 68.2 & 83\% & 1552.4 & \textbf{70.6} & 35.0 & 39.6 & \textbf{71.9} & 65.1 & 66.9 & 61.0 & 99.5 \\
   12 & 61.4 & 75\%  & 1568.5 & 70.1 & 34.8 & 39.8 & \textbf{71.9} & 65.0 & 66.7 & 61.0 & 99.5 \\
    16 & 54.6 & 67\% & 1561.0 & 70.1 & 35.0 & 39.7 & \textbf{71.9}  & 64.8  & 66.9 & 60.9 & 99.3 \\
    20 & 47.8 & 58\% & 1565.8 & 70.0 & 35.8 & \textbf{40.2} & 71.8 & 64.1 & \textbf{68.3} & 61.2 & 99.8 \\
    24 & 41.0 & 50\% & 1553.0 & 70.2 & \textbf{36.2} & 39.9 & 71.8  & 63.6 & 67.5 & 61.0 & 99.5 \\ 
    28 & 34.3 & 42\% & 1536.1 & 69.3 & 35.1 & 39.4 & 71.0 & 62.8 & 66.3 & 60.1 & 98.0 \\ 
    32 & 27.5 & 34\% & 1468.4 & 65.8 & 35.2 & 38.9 & 69.3 & 60.5 & 58.5 & 57.4 & 93.6 \\
    
    \bottomrule
  \end{tabular}%
  }
  \caption{Performance \vs Efficiency Balance of ShortV under different configurations.
\# ShortV Layers ($N$): the number of ShortV layers, Avg.: a normalized average score on benchmarks, Per.: the relative performance retention compared with the vanilla models.
  }
  \label{tab:next}
\end{table*}

In this section, we provide comprehensive ShortV performance under different settings, \ie different numbers of replaced layers.
The results are shown in Table~\ref{tab:next}.

\newpage

\end{document}